\documentclass{article}
\usepackage[preprint]{neurips_2025}

\usepackage{graphicx}       % 支持 \resizebox 等图形相关命令
\usepackage{booktabs}       % 支持 \toprule, \midrule, \bottomrule 等表格线条
\usepackage{multirow}       % 支持 \multirow 命令
\usepackage{array}          % 支持 >{\centering\arraybackslash}p{...} 这样的列格式
\usepackage[table,xcdraw]{xcolor} % （可选）如果你用了颜色，比如 \cellcolor，就需要这个包
\usepackage{caption}        % 支持自定义 caption 格式
\usepackage[utf8]{inputenc} % allow utf-8 input
\usepackage[T1]{fontenc}    % use 8-bit T1 fonts
\usepackage{hyperref}       % hyperlinks
\usepackage{url}            % simple URL typesetting
\usepackage{booktabs}       % professional-quality tables
\usepackage{amsfonts}       % blackboard math symbols
\usepackage{amsmath}        % for \argmin and other math operators
\usepackage{nicefrac}       % compact symbols for 1/2, etc.
\usepackage{microtype}      % microtypography
\usepackage{xcolor}         % colors
\usepackage{wrapfig}

\usepackage{amsmath}

\title{Unposed 3DGS Reconstruction with Probabilistic Procrustes Mapping}

\author{%
  Chong Cheng\thanks{Equal contribution.} \quad Zijian Wang$^*$ \quad  Sicheng Yu \quad Yu Hu \quad Nanjie Yao \quad Hao Wang\thanks{Corresponding author.}\\
  The Hong Kong University of Science and Technology (Guangzhou) \\
  \texttt{\{ccheng735, zwang886, yhu847\}@connect.hkust-gz.edu.cn} \\
  \texttt{yusch@mail2.sysu.edu.cn} \quad \texttt{nanjiey@uci.edu} \quad \texttt{haowang@hkust-gz.edu.cn}
}

\begin{document}

\maketitle

\begin{abstract}
% 3D Gaussian Splatting (3DGS) has emerged as a core technique for 3D representation, but its performance relies heavily on accurate camera poses and point cloud initialization, both of which are difficult to obtain in outdoor scenes. 
3D Gaussian Splatting (3DGS) has emerged as a core technique for 3D representation. 
% Its performance mainly relies on accurate camera poses and point cloud initialization, which can be obtained from pretrained Multi-View Stereo (MVS) models.
Its effectiveness largely depends on precise camera poses and accurate point cloud initialization, which are often derived from pretrained Multi-View Stereo (MVS) models. 
% Although pretrained Multi-View Stereo (MVS) models can estimate poses and geometry from unposed image sets, their scalability remains limited by memory and input size. 
However, in unposed reconstruction task from hundreds of outdoor images, existing MVS models may struggle with memory limits and lose accuracy as the number of input images grows. 
% To address this, we propose \textbf{DC-3DGS}, a scalable unposed 3DGS reconstruction framework that integrates pretrained MVS priors with a divide-and-conquer strategy. 
To address this limitation, we propose a novel unposed 3DGS reconstruction framework that integrates pretrained MVS priors with the probabilistic Procrustes mapping strategy.
% The method partitions input images into subsets, aligns the inferred submaps into a unified global coordinate system via a confidence-aware correspondence mapping module, and jointly refines scene geometry and camera poses using 3DGS. 
The method partitions input images into subsets, maps submaps into a global space, and jointly optimizes geometry and poses with 3DGS.
% Technically, we design a confidence-aware correspondence mapping module, formulate the mapping of tens of millions of point clouds as a Procrustes problem. 
% Technically, we design a confidence-aware correspondence mapping module and formulate the mapping of tens of millions of point clouds as a Procrustes problem. By solving a closed-form alignment with probabilistic coupling, our method maps globally point clouds and poses within minutes across hundreds of images. 
Technically, we formulate the mapping of tens of millions of point clouds as a probabilistic Procrustes problem and solve a closed-form alignment. By 
% solving a closed-form alignment with probabilistic coupling and a soft dustbin to reject uncertain correspondence, 
employing probabilistic coupling along with a soft dustbin mechanism to reject uncertain correspondences,
our method globally aligns point clouds and poses within minutes across hundreds of images.
% Additionally, we introduce an anchor-guided joint optimization pipeline for 3DGS and camera poses. Anchors are initialized from confidence-guided correspondences to construct the initial 3D Gaussians, which are then jointly optimized with camera poses through a differentiable 3DGS rendering pipeline. This further improves geometric accuracy and novel view synthesis quality. 
Moreover, we propose a joint optimization framework for 3DGS and camera poses. It constructs Gaussians from confidence-aware anchor points and integrates 3DGS differentiable rendering with an analytical Jacobian to jointly refine scene and poses, enabling accurate reconstruction and pose estimation.
Experiments on Waymo and KITTI datasets show that our method achieves accurate reconstruction from unposed image sequences, setting a new state of the art for unposed 3DGS reconstruction.
%without external pose priors.
\end{abstract}

\section{Introduction}

3D Gaussian Splatting (3DGS) has emerged as a revolutionary technique for 3D representation and novel view synthesis, owing to its superior rendering quality and real-time performance \citep{kerbl2023-3dgs, lu2024scaffold-gs}. By optimizing a set of 3D Gaussian parameters to represent the scene, 3DGS achieves high-fidelity and efficient visual effects, and has quickly become a focal point of research.

However, applying 3DGS to real-world scenarios, particularly for reconstruction from hundreds of uncalibrated outdoor images, remains highly challenging. Traditional 3DGS pipelines heavily rely on accurate precomputed camera poses and an initial point cloud \citep{schonberger2016sfm, schoenberger2016mvs}. These prerequisites are often difficult to obtain in complex outdoor environments, which significantly limits the broader applicability of 3DGS.

Several studies \citep{fu2024colmap-free3dgs,jiang2024COGS, dong2025Rob-GS, shi2025npnp} attempt to jointly optimize camera poses and Gaussian parameters from images in an end-to-end manner, thereby enabling unposed 3DGS reconstruction. 
However, they struggle on outdoor scenes due to scale ambiguity, sparse supervision, and sensitivity to noisy initialization, often resulting in limited accuracy \citep{fan2024instantsplat}. 
Another common strategy combines Structure-from-Motion (SfM) with 3DGS \citep{schonberger2016sfm}, but the SfM phase is typically computationally expensive, often requiring hours of processing and being prone to failure in challenging outdoor conditions.

% Recently, pretrained Multi-View Stereo (MVS) models have gained traction as a promising alternative, since they are capable of inferring dense point clouds and camera poses directly from images. Some feed-forward 3DGS methods predict Gaussians directly from images but typically support only a dozen views, limiting their applicability to large-scale scenes. However, existing methods are often memory-intensive and limited to sparse-view reconstruction due to frequent OOM issues. Recent pretrained MVS models support larger image batches, but still suffer from accuracy degradation and memory bottlenecks as input size grows.
Pretrained Multi-View Stereo (MVS) models \citep{wang2024dust3r, leroy2024mast3r} have long served as a structured approach for inferring dense point clouds and camera poses directly from images, and remain a promising foundation for unposed 3D reconstruction. In contrast, feed-forward 3DGS methods predict Gaussians directly from images with improved efficiency, but typically support only a dozen views and are prone to out-of-memory (OOM) issues \cite{ye2024noposplat,xu2024freesplatter,chen2024pref3r,zhang2025flare}. While modern MVS models can handle larger input batches, they still face accuracy degradation and memory bottlenecks as the number of views increases, especially in outdoor scenes \citep{wang2025vggt,Yang_2025_fast3R}.
% However, such approaches tend to be inefficient and memory-intensive. They frequently result in out-of-memory (OOM) errors, which restricts their use to sparse-view reconstructions. More recent pretrained MVS models aim to infer both camera poses and dense geometry from unordered image collections. While these models are more scalable, their performance tends to degrade significantly as the number of input images increases, and they still suffer from memory limitations.

% These challenges naturally motivate the use of a divide-and-conquer strategy. This involves decomposing a large scene into multiple independent sub-scenes, processing each one individually, and then merging them into a globally consistent reconstruction. However, because each submap is inferred in its own local coordinate frame, the resulting reconstructions often suffer from scale ambiguity and geometric inconsistencies. Existing registration methods, as well as direct relative pose estimation using pretrained models, typically fail to provide accurate alignment when faced with variation and noise. Therefore, a core challenge in this setting is how to efficiently and robustly align independently inferred submaps into a globally consistent coordinate system, so as to enable high-quality 3DGS reconstruction.
These challenges inspire a divide-and-conquer strategy: decomposing a large image collection into smaller subsets, processing them individually, and merging them into a globally consistent reconstruction. However, since each submap is inferred in its own local frame, the results often suffer from scale ambiguity and geometric inconsistency. Existing registration methods \citep{yang2020teaserfastcertifiablepoint, besl1992ICP, Lawrence_2019, chen2024pointreggptboosting3dpoint} typically fail under scale ambiguity, geometric deviations, and the computational challenge of aligning tens of millions of points. A key challenge, therefore, is how to efficiently align these submaps into a unified coordinate system to enable high-quality 3DGS reconstruction.

% To address these challenges, we propose \textbf{DC-3DGS}—a collaborative framework for unposed 3DGS reconstruction that combines pretrained MVS priors with a divide-and-conquer strategy. The proposed method combines the feed-forward MVS models' priors and overlapping information across image groups, progressively recovers globally consistent point clouds and camera poses from local submaps, leading to high-quality 3DGS reconstruction.
To address these challenges, we propose a collaborative framework for unposed 3DGS reconstruction that integrates pretrained MVS with a divide-and-conquer strategy. Using feed-forward priors and overlapping views across image groups, we progressively recover globally consistent point clouds and camera poses from local submaps, leading to high-quality 3DGS reconstruction.

% Specifically, we formulate submap mapping as a Procrustes problem by constructing confidence-aware correspondences between overlapping frames. 
% Specifically, we design a correspondence mapping strategy that constructs reliable associations from overlapping frames, structurally transforming the problem of aligning tens of millions of point clouds into a Procrustes formulation. 
Specifically, we reformulate the original alignment of tens of millions of points as a probabilistic Procrustes problem by designing overlapping-frame correspondences at the pixel level. We first obtain a closed-form $Sim(3)$ solution using the Kabsch-Umeyama algorithm, and then refine it via a probabilistic coupling with a soft dustbin mechanism that rejects uncertain matches. This approach effectively resolves scale ambiguity and local geometric discrepancies between submaps, achieving robust global alignment 
within minutes
% with $\mathcal{O}(N \log N)$ complexity.

% Further, we introduce a confidence-aware correspondences joint optimization framework that integrates high-confidence correspondences to construct a globally consistent 3D Gaussian representation. By combining a differentiable 3DGS rendering pipeline with photometric and geometric constraints, we jointly optimize camera poses and 3D Gaussian scene, leading to significantly improved pose accuracy and view synthesis quality.

Further, we propose a joint optimization framework for 3DGS and camera poses, where Gaussians are initialized from downsampled anchor points obtained via confidence-aware correspondence filtering. Camera poses are optimized through differentiable 3DGS rendering, with gradients propagated via an analytical quaternion Jacobian, leading to improved pose accuracy and view synthesis quality.

\noindent{Our main contributions are as follows:}
\begin{enumerate}
\item We propose an alignment method that casts submap mapping as a probabilistic Procrustes problem. It combines closed-form $Sim(3)$ estimation with probabilistic and outlier rejection, enabling global pose and point-cloud recovery from hundreds of images within minutes.
% We propose a hybrid probabilistic Procrustes alignment method based on confidence-aware overlapping correspondences. It efficiently and accurately aligns feed-forward MVS submaps, enabling global pose and point-cloud recovery from hundreds of unposed images in minutes.
\item 
We propose a 3DGS and pose joint optimization module that constructs Gaussians from confidence-guided anchor points and refines scene and poses via 3DGS differentiable rendering with an analytical Jacobian, improving pose accuracy and reconstruction quality.
% We introduce a confidence-aware correspondences joint 3DGS optimization scheme that uses anchor points and a differentiable 3DGS rendering pipeline to jointly optimize camera poses and 3D Gaussian scene representation, enhancing both pose accuracy and reconstruction fidelity.
\item Experiments on the Waymo and KITTI datasets demonstrate that our method achieves highly efficient and accurate global reconstruction from unposed images, setting a new state of the art for unposed 3DGS reconstruction.
\end{enumerate}

\section{Related Work}

\subsection{Unposed 3D Gaussian splatting}
% 3D Gaussian Splatting (3DGS) \citep{kerbl2023-3dgs} traditionally depends on accurate camera poses and sparse point clouds from COLMAP \citep{schonberger2016sfm}. 
Traditional 3D Gaussian Splatting (3DGS) \citep{kerbl2023-3dgs} relies on accurate camera poses and sparse point clouds typically provided by COLMAP \citep{schonberger2016sfm}. Due to COLMAP's high computational cost and limited robustness in challenging conditions, recent works aim to recover camera parameters and reconstruct Gaussian scenes directly from multi-view images.

%Several 3DGS-based SLAM methods \citep{matsuki2024monogs, huang2024photo-slam, yu2025opengs-slam} adopt differentiable rendering pipelines to jointly optimize camera poses and scene representations. 
CF-3DGS \citep{fu2024colmap-free3dgs} initializes the Gaussian field using monocular depth and progressively refines both camera parameters and Gaussians to support unposed reconstruction. COGS \citep{jiang2024COGS} incrementally reconstructs the scene by registering cameras through 2D correspondences, while Rob-GS \citep{dong2025Rob-GS} introduces a robust pairwise pose estimation strategy. NoParameters \citep{shi2025npnp} jointly optimizes intrinsics, extrinsics, and Gaussians, removing the need for prior camera calibration. InstantSplat \citep{fan2024instantsplat} leverages the pre-trained pointmap model DUSt3R \citep{wang2024dust3r} for initialization and accelerates optimization via parallel grid partitioning, but remains limited to sparse-view scenarios with relatively few images.

% Another line of work \cite{smart2024splatt3r,ye2024noposplat,charatan2024pixelsplat} leverages pre-training to enable feed-forward networks that directly predict high-quality Gaussian scenes from paired input images. Extensions of this approach \cite{xu2024freesplatter,chen2024pref3r,zhang2025flare} extend this idea to support more input images and improved reconstruction quality. However, as scene size and view count increase, these approaches face significant memory and runtime demands, or exhibit degraded reconstruction robustness.
Another line of work \cite{smart2024splatt3r,ye2024noposplat,charatan2024pixelsplat} leverages pre-training to enable feed-forward networks that directly predict high-quality Gaussian scenes from paired images. Recent extensions \cite{xu2024freesplatter,chen2024pref3r,zhang2025flare} support more inputs and improve quality, but typically scale only to a dozen views. As scene size and view count grow, these methods face significant memory and runtime demands or degraded robustness.

% In contrast, our method handles pose-free outdoor scenes with large numbers of input views, achieving scalable and accurate reconstruction with practical memory and runtime requirements.
To enable scalable unposed 3DGS on outdoor scenes with hundreds of images, we introduce pretrained MVS models and adopt a divide-and-conquer strategy to 3DGS reconstruction.
%In contrast, our method efficiently handles unordered, unposed images, achieving accurate reconstruction and pose estimation at scale with practical resource usage.
%In contrast, our method supports unordered image collections without any known pose. It maintains high reconstruction and pose estimation accuracy in large-scale, densely-viewed scenes, while keeping memory and runtime requirements within practical bounds.

\subsection{Multi-view 3D Reconstruction}
Traditional multi-view reconstruction pipelines \citep{hartley2003mvg} consist of handcrafted stages including feature matching, triangulation, and bundle adjustment. Systems like COLMAP \citep{schonberger2016sfm,mur2017orb-slam2, schoenberger2016mvs} perform well in static scenes, but suffer from accumulated errors, high computational cost, and failure in challenging scenarios.

Learning-based approaches \citep{yao2018mvsnet,yao2019R-mvsnet,zhang2023geomvsnet,ma2022CER-MVS} leverage end-to-end networks to recover high-quality geometry from calibrated images. More recently, end-to-end differentiable SfM frameworks \citep{wei2020deepsfm,wang2021deep-2vsfm,wang2024vgg-sfm,smith25flowmap} aim to jointly estimate camera parameters and scene structure directly from image collections.

DUSt3R \citep{wang2024dust3r} and MASt3R \citep{leroy2024mast3r} regress dense point clouds and camera parameters from paired images, replacing handcrafted components with pre-trained backbones. This feedforward paradigm has been extended to multi-image settings using memory encoders \citep{wang2024spann3r, wang2025CUT3R, cabon2025must3r} and subgraph fusion networks \citep{liu2024slam3r}. VGGT \citep{wang2025vggt} and Fast3R \citep{Yang_2025_fast3R} further adopt global attention mechanisms to reason across multiple views. MV-DUSt3R+ \citep{tang2024mv-dust3r+} and FLARE \citep{zhang2025flare} enable end-to-end 3D Gaussian Splatting reconstruction from sparse-view inputs, and similar strategies have been applied to dynamic scene modeling \citep{zhang2024monst3r, chen2025easi3r}. However, these methods struggle with increasing view counts, facing memory bottlenecks and degraded reconstruction robustness. Inconsistent in structure and scale across independently processed submaps complicates global alignment, limiting fidelity in large-scale outdoor scenes.

% We adopt a divide-and-conquer strategy to partition large scenes into multiple independent sub-scenes, and propose a hybrid probabilistic Procrustes alignment method for robust global alignment. Building on this, we perform anchor-based joint 3DGS optimization, enabling efficient and accurate reconstruction from multi-view inputs.
To address these challenges, we adopt a divide-and-conquer strategy that partitions images into local submaps and reconstructs a globally consistent 3DGS scene through alignment and joint optimization.

%%%%%%%%%%%%%%%%%%%%%%%%%%%%%%%%%%%%%%%%%%%%%%%%%%%%%%%%%%%%%%%%%%%%%%%%%%%%%%%%%%%%%%%%%%%%%%%%%%%%%%%%%%%%%%%%%%%%%%%%%%%%%%%%%%%%%%%%%%%%%%%%%%%%%%%%%%%%%%%%%%%%%%%%%%%%%%%%%%%%%%%%%%%%%%%%%%%%%%%%%%%%%%%%%%%%%%%%%%%%%%%%%%%%%%%%%%%%%%%%%%%%
%%%%%%%%%%%%%%%%%%%%Methods%%%%%%%%%%%%%%%%%%%%%%%%%%%%%%%%%%%%%%%%%%%%%%%%%%%%%%%%%%%%%%%%%%%
\section{Method}
\begin{figure*}
  \centering
  \includegraphics[width=\textwidth]{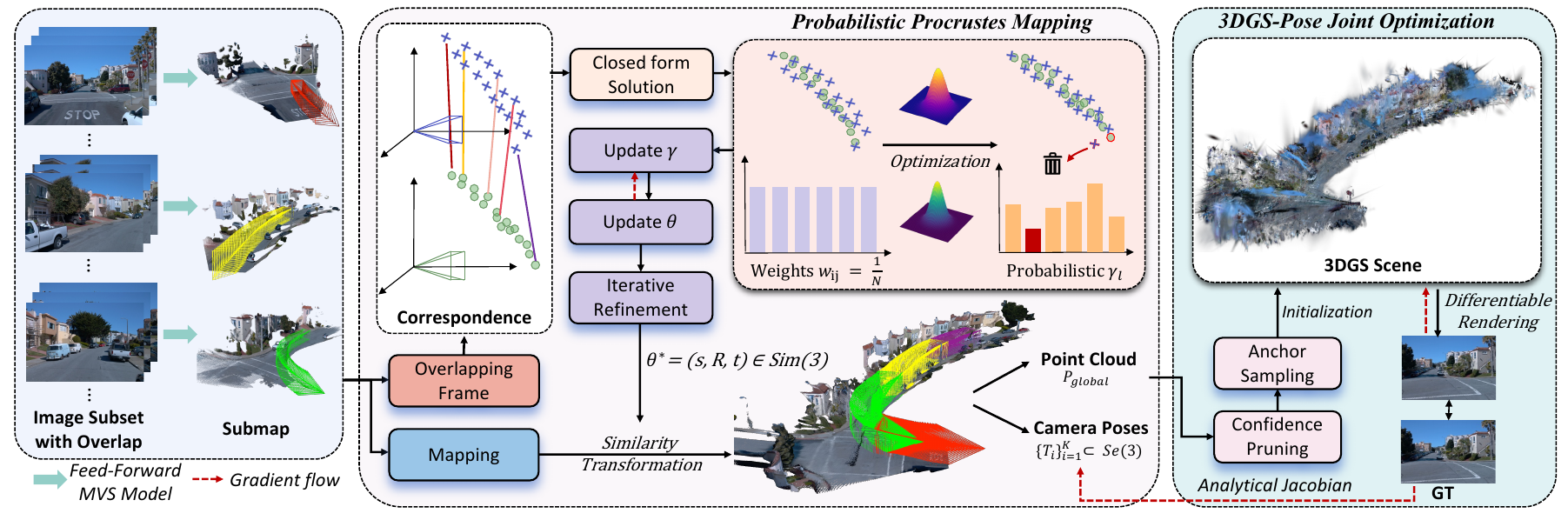}
  \vspace{-13pt}
  \caption{
% Pipeline of our proposed method.
% We first partition the unposed image sequence into multiple subsets and apply a pretrained MVS model to infer local point clouds and camera poses. We design overlapping-frame correspondences to convert large-scale submap alignment into a Procrustes problem, solved by a closed-form Sim(3) estimator and probabilistic refinement. 
% Final 3DGS and poses are jointly optimized using anchor-based initialization and differentiable rendering with analytical Jacobians.
We begin by partitioning the unposed image sequence into multiple subsets, and apply a pretrained MVS model to infer local point clouds and camera poses. Overlapping-frame correspondences are constructed to reformulate large-scale submap alignment as a probabilistic Procrustes problem. This is solved via a closed-form $Sim(3)$ estimator, followed by probabilistic refinement and soft outlier rejection.
The final 3DGS and poses are jointly optimized through anchor-based initialization and differentiable rendering, with gradients propagated via analytical Jacobians.
} 
  \label{fig:framework}
  \vspace{-13pt}
\end{figure*}

We aim to reconstruct high-quality 3D Gaussian scenes from hundreds of unposed outdoor images.
As illustrated in Fig.~\ref{fig:framework}, the image set is partitioned into overlapping subsets, each independently processed by a pretrained MVS model to estimate local point clouds and camera poses.
These submaps are then globally aligned via probabilistic Procrustes mapping, followed by joint optimization of the 3DGS and camera poses, resulting in high-fidelity and globally consistent reconstructions.

\subsection{Problem Formulation} 
Given \(K\) images \(\{I_k\}_{k=1}^{K}\) split into \(G\) fixed‐size subsets$\{\mathcal{S}_g\}_{g=1}^{G}$, each subset is fed to a pretrained MVS network to produce a local submap $\mathcal M_g=(\mathbf P_g,\{T_i^{(g)}\}_{i\in\mathcal S_g})$,
containing a dense point cloud and its camera poses. Our goal is to fuse these into a globally consistent scene \((\mathbf P_{\mathrm{global}},\{T_i\}_{i=1}^K)\).

To achieve globally consistent alignment across submaps, we aim to estimate the optimal similarity transformation $\theta = (s, R, \mathbf{t}) \in \mathrm{Sim}(3)$ between submaps, where $s \in \mathbb{R}^+$ is the scale factor, $R \in SO(3)$ is the rotation matrix, and $\mathbf{t} \in \mathbb{R}^3$ is the translation vector. However, feed‐forward MVS submaps suffer from scale ambiguity and geometric distortions. The structural complexity of outdoor scenes further leads to the failure of standard registration methods. Moreover, each submap typically contains tens of millions of points, making global alignment a high-dimensional and computationally intensive task that challenges both accuracy and efficiency.

To address these challenges, we define $k$ overlapping frames between each pair of adjacent subsets, denoted as $\mathcal{O}_{ab}$. This allows us to reformulate multi-submap alignment as a classical Procrustes problem. We then extract per-pixel 3D correspondences between submaps \(a\) and \(b\) within $\mathcal{O}_{ab}$:

\begin{equation}
    \mathcal C_{ab}=\bigl\{(\mathbf p_i^a,\mathbf q_j^b)\mid
    \pi\bigl(T_i^a\mathbf p_i^a\bigr)=\pi\bigl(T_j^b\mathbf q_j^b\bigr)\bigr\},
\end{equation}

where \(\pi:\mathbb R^3\to\mathbb R^2\) is the standard camera projection model. We then solve the classical Procrustes problem as an optimization that minimizes the distances between transformed point pairs:
\begin{equation}
\label{eq:procrustes}
  \theta^*=\underset{s>0,R\in SO(3),\mathbf t\in\mathbb R^3}{\arg\min}
    \sum_{(i,j)\in\mathcal C_{ab}}
    \bigl\|s\,R\,\mathbf p_i^a+\mathbf t-\mathbf q_j^b\bigr\|^2.
\end{equation}

\subsection{Probabilistic Procrustes Mapping}
\subsubsection{Procrustes Closed-form Solution} 

To efficiently estimate the optimal similarity transformation in Eq.~\eqref{eq:procrustes}, we adopt the Kabsch-Umeyama algorithm \citep{88573,Lawrence_2019} to compute a closed-form solution based on the correspondence set $\mathcal{C}_{ab}$. First, we compute the centroids of each point set:
\begin{equation}
\bar{\mathbf{p}} = \frac{1}{N}\sum\nolimits_{(i,j)\in\mathcal{C}_{ab}}\mathbf{p}_{i}^{\,a},\qquad
\bar{\mathbf{q}} = \frac{1}{N}\sum\nolimits_{(i,j)\in\mathcal{C}_{ab}}\mathbf{q}_{j}^{\,b},
\end{equation}
where $N = |\mathcal{C}_{ab}|$ denotes the number of point pairs. These centroids reflect the global offsets of the two point clouds and will be used to compute the translation vector. Next, we construct the cross-covariance matrix $\Sigma$ between the two point clouds to capture their spatial correlation structure:
\begin{equation}
\Sigma = \frac{1}{N}\sum_{(i,j)\in\mathcal{C}_{ab}}
(\mathbf{p}_{i}^{\,a}-\bar{\mathbf{p}})
(\mathbf{q}_{j}^{\,b}-\bar{\mathbf{q}})^{\!\top}.
\end{equation}
By performing singular value decomposition (SVD) $\Sigma = U\Lambda V^{\top}$, we obtain the principal directions of the two point sets. This allows us to compute the closed-form solution of $\theta^*$:

\begin{equation}
R_0 = U\,\operatorname{diag}\!\bigl(1,\,1,\,\det(UV^{\top})\bigr)\,V^{\top},\quad
s_0 = \frac{\operatorname{tr}(\Lambda)}{\operatorname{tr}(\Sigma_{p})}, \quad
\mathbf{t}_0 = \bar{\mathbf{q}} - s_0 R_0 \bar{\mathbf{p}},
\end{equation}
where $R_0$ is a proper rotation ensuring a right-handed coordinate system, $s_0$ is given by the ratio between the singular values and the variance of the source point cloud, and $\mathbf{t}_0$ aligns centroids. This closed‐form step provides an efficient initialization for refining submap alignment.

Although the closed‐form solution is theoretically optimal, it relies on two critical assumptions: 
\begin{enumerate}
    \item The spatial distributions of the two point sets must be identical.%, i.e., $\frac{1}{N}\sum\mathbf{p}_{i}^{\,a}\mathbf{p}_{i}^{\,a\top}=\frac{1}{N}\sum\mathbf{q}_{j}^{\,b}\mathbf{q}_{j}^{\,b\top}$.
    \item The correspondences must be noise-free, i.e., $\bigl\|s_0\,R_0\,\mathbf p_i^a+\mathbf t_0-\mathbf q_j^b\bigr\|=0,\;\forall(i,j)\in\mathcal{C}_{ab}$.
\end{enumerate}

These conditions hold in ideal cases where point clouds are perfectly accurate and geometrically consistent. However, in practical monocular reconstruction, even with overlapping frames providing pixel-level correspondences, the spatial distributions of the 3D points may vary significantly. This violates the assumptions of the closed-form solution and often leads to systematic bias when used directly, making the mapping results less robust.

\subsubsection{Probabilistic Mapping}

We observe that point clouds predicted by feed-forward MVS models exhibit structural bias due to learned priors, which leads to systematic errors in closed-form alignment. To address this, we formulate point cloud registration as a probabilistic Procrustes problem augmented with a dustbin mechanism. Specifically, we associate each candidate correspondence \((\mathbf{p}_l, \mathbf{q}_l)\) between submaps with a probabilistic matching weight \(\gamma_l \in [0,1]\).

To handle outliers, we introduce a probability-based \emph{dustbin} mechanism: a control parameter \(\eta \in [0,1]\) specifies the maximum allowable fraction of correspondences that can be excluded from alignment. To implement this, we augment the target set with a virtual dustbin point \(\mathbf{q}_{\text{dustbin}}\), and assign it a fixed marginal weight \(b_{\text{dustbin}} = \delta\).

Our objective is to jointly optimize the similarity transformation \(\theta = (s, R, \mathbf{t}) \in \mathrm{Sim}(3)\) and the correspondence probabilities \(\{\gamma_l\}_{l=1}^N\). Each weight \(\gamma_l\) encodes the soft association strength between a source point \(\mathbf{p}_l\) and its target \(\mathbf{q}_l\) under the current transformation. The objective is formulated as:

\begin{equation}
\min_{s, R, \mathbf{t}, \gamma} \sum_{l} \gamma_{l} \| s R \mathbf{p}_l + \mathbf{t} - \mathbf{q}_l \|^2 + \epsilon \sum_{l} \gamma_{l} \ln \gamma_{l}, \quad \text{subject to} \quad \sum_{l} \gamma_{l} = 1, 
\end{equation}

where $\gamma_{l} \in [0,1]$ denotes the soft matching probability between source point $\mathbf{p}_l$ and target point $\mathbf{q}_l$.

\paragraph{Probability Weights Update.}
We initialize the transformation parameters \(\theta^{(0)} = (s^{(0)}, R^{(0)}, \mathbf{t}^{(0)})\) using the closed-form Kabsch–Umeyama method. Given a fixed transformation, the correspondence weights \(\gamma\) are updated via entropy-regularized optimization:

\begin{equation}
\gamma_{l} \propto \exp\left( -\frac{\| s R \mathbf{p}_l + \mathbf{t} - \mathbf{q}_l \|^2}{\epsilon} \right),
\end{equation}
where the proportionality is followed by a normalization step to satisfy the marginal constraints using the step-wise iteration optimization. 

\paragraph{Transformation Update.}

Fixing \(\gamma\), we compute the gradients of the objective \(\mathcal{L}_\theta\) with respect to the transformation parameters \(\theta = (s, R, \mathbf{t})\). Let \(\mathbf{p}_l' = R\mathbf{p}_l\), Then:

\begin{align}
\nabla_{\mathbf{t}}\mathcal{L}_\theta &= 2 \sum_{l=1}^{N} \gamma_{l}^{(k)} \left( s \mathbf{p}_{l}' + \mathbf{t} - \mathbf{q}_{l} \right).\\
\nabla_{s}\mathcal{L}_\theta &= 2 \sum_{l=1}^{N} \gamma_{l}^{(k)} \left( s \mathbf{p}_{l}' + \mathbf{t} - \mathbf{q}_{l} \right)^{\top} \mathbf{p}_{l}'.
\end{align}

To update the rotation $R$, we parameterize it using a unit quaternion $q=(w,v^\top)^\top, \text{where} \ v=(x,y,z)^\top$, and use the chain rule to compute:

\begin{equation}
\nabla_{q} \mathcal{L}_\theta = 2 \sum_{l=1}^{N} \gamma_{l}^{(k)} \left( s R(q)\mathbf{p}_l + \mathbf{t} - \mathbf{q}_l \right)^{\top} s \frac{\partial (R(q)\mathbf{p}_l)}{\partial q}.
\end{equation}

The transformation parameters are updated using gradient descent:
\begin{equation}
    \theta^{(k+1)} = \theta^{(k)} - \eta_\theta \nabla_\theta \mathcal{L}_\theta(\theta^{(k)}),
\end{equation}

where \(\eta_\theta\) is the learning rate. The optimization terminates when either the pose converges or a maximum number of iterations is reached. In practice, the accurate closed-form initialization typically leads to convergence within a few iterations.

The resulting optimal transformation \(\theta^\star = (s_g, R_g, \mathbf{t}_g)\) is then applied to all 3D points and camera poses in submap \(\mathcal{S}_g\), transforming them into the global coordinate frame and updating the corresponding poses. Iteratively applying this procedure across all submaps yields a globally consistent point cloud and a unified camera trajectory.

\subsection{3DGS and Pose Joint  Optimization}
\label{sec:anchor_optim}
After submap alignment, we obtain initial camera poses and a dense point cloud in a unified global coordinate system. However, due to the inherent scale uncertainty, depth noise, and residual pose drift in monocular reconstruction, further refinement is necessary. 
To this end, we jointly optimize 3D Gaussian parameters and camera poses using a differentiable 3DGS rendering pipeline, improving both pose accuracy and reconstruction quality.

\paragraph{3D Gaussian Splatting.}
\label{3dgs}
We model the scene as a set of 3D Gaussians: $\mathcal{G} = \{\mathcal{G}_i:(\boldsymbol{\mu}_i, \boldsymbol{\Sigma}_i, \mathbf{c}_i, \Lambda_i)|i=1,...,N\}$. Each Gaussian point is defined by position $\mu_i$, 3D covariance matrix $\Sigma_i\in\mathbb{R}^{3\times 3}$, opacity $\Lambda_i$ and color $\mathbf{c}_i$ obtained by spherical harmonics.

For a specific view, given camera pose $T=(R,t)$ and camera intrinsic $\mathbf{K}\in \mathbb{R}^{3\times3}$, we can the render RGB image $\hat{I}$ via rasterization pipeline. First, project our 3D Gaussians to 2D image plane:
\begin{equation}
    \mu'=\pi(T\cdot \mu), \qquad \Sigma' = JW\Sigma W^\top J^\top,
\end{equation}
where $\pi$ is the projection operation, $W$ is the rotational component of $T$, and $J$ is the Jacobian of the affine approximation of the projective transformation. Then, the color of pixel can be formulated as the alpha-blending of $N$ ordered points that overlap the pixel:
\begin{equation}
    C=\sum_{i\in N}c_i \alpha_i \prod_{j=1}^{i-1}(1-\alpha_j),
\end{equation}
where $\alpha_i$ is the density given by evaluating a 2D Gaussian with covariance $\Sigma'$.

\paragraph{Joint Optimization.}

We first extract a high-confidence subset from the global point cloud and apply downsampling to obtain the initial anchor set $\mathcal{A} = \{\mathbf{x}_i\}$, which is used to initialize 3D Gaussian $\mathcal{G}_i = (\boldsymbol{\mu}_i, \boldsymbol{\Sigma}_i, \mathbf{c}_i, \Lambda_i)$. These Gaussians $\mathcal{G}=\{\mathcal{G}_i\}$ together form the initial global scene.

Based on this, we define a closed-loop optimization framework that jointly optimizes the camera poses $T = (R, \mathbf{t})$ and Gaussian parameters $\{\boldsymbol{\mu}_i, \boldsymbol{\Sigma}_i, \mathbf{c}_i, \Lambda_i\}_{i=1}^N$ through the following objective:
\begin{equation}
\label{eq:joint_obj}
\mathcal{L}_{\mathrm{total}}
=\alpha\,
\lVert \hat I_k - I_k \rVert_1
+
(1 - \alpha)\text{SSIM}(\hat I_k, I_k),
\end{equation}
where $\hat I_k$ denotes the rendered image under the current view $k$, $I_k$ is the ground truth, and $\alpha$ is the weight balancing the L1 and SSIM terms.

From the 3DGS rendering pipeline, it follows that the gradient of the camera pose \(T\) depends on two intermediate quantities: \(\Sigma'\), and the projected coordinates \(\mu'_i\) of each Gaussian \(\mathcal{G}_i\). By applying the chain rule, we can derive a fully analytic expression for $\frac{\partial \mathcal{L}}{\partial T}$, thereby avoiding the runtime overhead of automatic differentiation and ensuring numerical stability during quaternion normalization. The resulting analytic gradient takes the following form:
\begin{equation}
\begin{aligned}
\frac{\partial \mathcal{L}}{\partial T}
&= \frac{\partial \mathcal{L}}{\partial \hat I_k}\,\frac{\partial \hat I_k}{\partial T}
%= \frac{\partial \mathcal{L}}{\partial \hat I_k}\Bigl(\frac{\partial \hat I_k}{\partial c_i}\,\frac{\partial c_i}{\partial T}
%+ \frac{\partial \hat I_k}{\partial \alpha_i}\,\frac{\partial \alpha_i}{\partial T}\Bigr)\\
= \frac{\partial \mathcal{L}}{\partial \hat I_k}\frac{\partial \hat{I}_k}{\partial \alpha_i}\Bigl(\frac{\partial \alpha_i}{\partial \Sigma'}\,\frac{\partial \Sigma'}{\partial T}
+ \frac{\partial \alpha_i}{\partial \mu'}\,\frac{\partial \mu'}{\partial T}\Bigr),
%&= \frac{\partial \mathcal{L}}{\partial \hat I_k}\frac{\partial \hat{I}_k}{\partial \alpha_i}\Bigl(\frac{\partial \alpha_i}{\partial \Sigma'}\,\frac{\partial \bigl(J\,T\,\Sigma\,T^{\top}J^{\top}\bigr)}{\partial T}
%+ \frac{\partial \alpha_i}{\partial m_i}\,\frac{\partial \bigl(K\,T\,\mu_i\bigr)}{\partial Td_i}\Bigr).
\end{aligned}
\end{equation}
\begin{equation}
    \frac{\partial \Sigma'}{\partial T} = \frac{\partial\Sigma'}{\partial J}\frac{\partial J}{\partial \mu_c}\frac{\partial \mu_c}{\partial T}+\frac{\partial \Sigma'}{\partial W}\frac{\partial W}{\partial T},
\end{equation}
\begin{equation}
    \frac{\partial \mu'}{\partial T}=\frac{\partial \mu'}{\partial \mu_c}\frac{\partial \mu_c}{\partial T},
\end{equation}
where $\mu_c$ denotes the point $\mu$ in world coordinates transformed into the camera frame by the pose $T$. We parameterize the camera pose $T=(R,t)$ by a unit quaternion $q\in\mathbb{R}^4$ and a translation vector $t\in\mathbb{R}^3$, and we provide the analytic gradients with respect to $q$ and $t$ in Appendix~\ref{jacob2}.

%\paragraph{Parameter Constraints and Updates.}
To keep $q$ unit-length, we apply the projected gradient updates:
\begin{equation}
    q \leftarrow \frac{q - \eta \nabla_q \mathcal{L}}{\lVert q - \eta \nabla_q \mathcal{L} \rVert}.
\end{equation}

%The translation vector $\mathbf{t}$ is updated via standard gradient descent. This update scheme ensures numerical stability and satisfies the physical constraints of the transformation.

By jointly optimizing the camera poses, 3D Gaussian parameters, and image reprojection, we obtain a globally consistent 3D Gaussian scene with accurate pose estimation and high-fidelity rendering.

\begin{figure*}
  \centering
  \includegraphics[width=\textwidth]{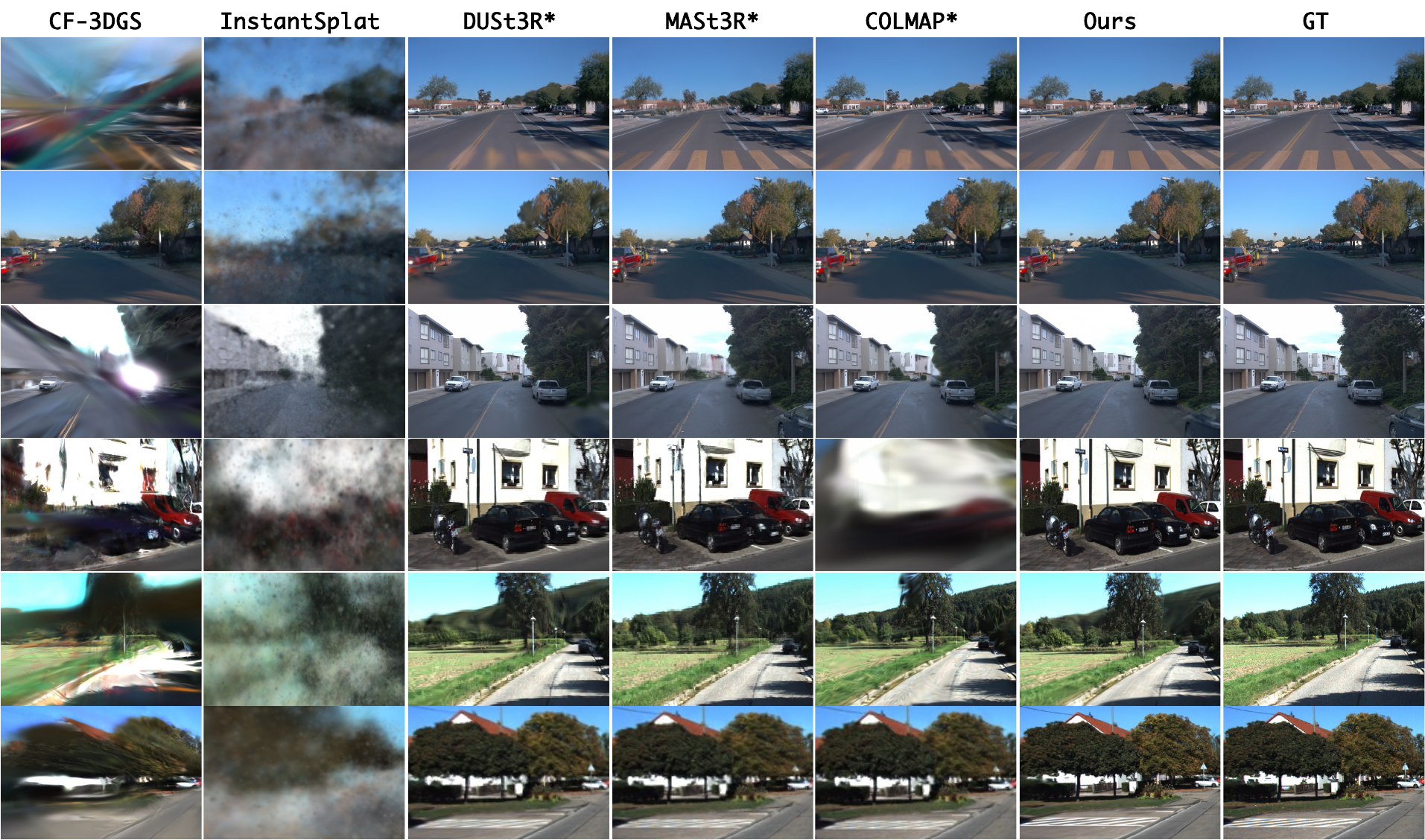}
  \vspace{-12pt}
  \caption{\textbf{Qualitative Comparison on Waymo (top three rows) and KITTI (bottom three rows).} Methods marked with an asterisk (*) are reconstructed using 3DGS. InstantSplat is trained on only 80 images due to memory constraints. Our method achieves high-fidelity image reconstruction with clearer textures and finer details.}%In contrast, GlORIE-SLAM and OpenGS-SLAM exhibit rendering distortions and blurriness.
  \label{fig:waymo_render}
  \vspace{-15pt}
\end{figure*}

\section{Experiments}
\subsection{Experimental Setup}
\textbf{Implementation details.} 

Our experiments are implemented using the PyTorch framework and conducted on a single NVIDIA RTX A6000 GPU with an AMD EPYC 7542 CPU. All results are reported using the best-performing pretrained MVS model, VGGT \citep{wang2025vggt}. 

We set the group size to 60 and the inter-group overlap to $K=1$, which empirically yielded the best performance.
Camera poses are optimized with an initial learning rate of $10^{-5}$, decayed to $10^{-7}$ until convergence. The dustbin capacity is set to 20\%. 
To enable efficient optimization, we first prune the lowest-confidence 3\% of points and then apply voxel-based downsampling to retain 0.05\% of points as anchors. Additional implementation details are provided in the supplementary material.

\textbf{Dataset.} We conduct evaluations on two outdoor datasets, selecting 9 scene groups from Waymo~\cite{Sun_2020_CVPRwaymo} and 8 from KITTI~\cite{geiger2013kittidata}, each consisting of 200 front-view images captured under diverse conditions. All images are used for evaluation. We assess both the image reconstruction quality and the accuracy of the estimated camera poses across entire sequences.

\begin{table*}
\centering
\caption{\textbf{Quantitative results on Waymo and KITTI datasets.} T$_\mathrm{m}$ denotes matching time and T$_\mathrm{t}$ denotes training time. Methods marked with an asterisk (*) indicate methods reconstructed using 3DGS. Flare fails due to out-of-memory (OOM). ATE measures pose accuracy; PSNR, SSIM, and LPIPS evaluate image reconstruction quality. Best results are in \textbf{bold}. Our method achieves the best accuracy and reconstruction fidelity.}
\vspace{-5pt}
\fontsize{7}{7}\selectfont  % 设置整体字体大小
\setlength{\tabcolsep}{2pt} % 控制列间距
\renewcommand{\arraystretch}{1.3} % 控制行高
    \begin{tabular}{@{}l|ccccccc|ccccccccc@{}}
    % {>{\centering\arraybackslash}p{2.2cm}|*{4}{>{\centering\arraybackslash}p{1.0cm}}*{2}{>{\centering\arraybackslash}p{0.6cm}}|>{\centering\arraybackslash}p{1.0cm}*{4}{>{\centering\arraybackslash}p{1.0cm}}>{\centering\arraybackslash}p{1.0cm}*{2}{>{\centering\arraybackslash}p{0.6cm}}}
    \toprule
    \multirow{2}{*}{\textbf{Method}} & \multicolumn{7}{c}{\textbf{Waymo} \citep{Sun_2020_CVPRwaymo}} & \multicolumn{7}{c}{\textbf{KITTI} \cite{geiger2013kittidata}}  \\ \cmidrule(lr){2-8} \cmidrule(lr){9-15}  
                        & ATE$\downarrow$ & PSNR$\uparrow$ & SSIM$\uparrow$ & LPIPS$\downarrow$ & $T_m$ & $ T_t $ & GPU & ATE$\downarrow$ & PSNR$\uparrow$ & SSIM$\uparrow$ & LPIPS$\downarrow$ & $T_m$ & $ T_t $  & GPU  \\ \midrule
    COLMAP+SPSG*  & 3.68 & 30.17 & 0.893 & 0.314 & 45min & 58min& 13GB & 12.1 & 19.52 & 0.647 & 0.438 & 41 min & 35 min & 10GB\\ 
    CF-3DGS    &  5.46   &  22.69   & 0.736   &  0.316  & - &67min &12GB  & 5.99   & 15.91   & 0.490   &  0.486  & - &195min  &11GB   \\ 
    DUSt3R*  & 5.59 & 29.39 & 0.871 & 0.310 & 38min &42min & 46GB & 3.10 & 23.87 & 0.767 & 0.311 & 32min &51min & 45GB\\ 
    MASt3R*  & 6.12 & 27.49 & 0.867 & 0.308 &82min &46min & 40GB & 5.71 & 23.63 & 0.778 & 0.274 
    &94min &57min & 37GB\\ 
    Fast3R*   & 43.9 & 20.66 & 0.764 & 0.468 & 30s &46min & 14GB & 47.3 & 16.73 & 0.533 & 0.581 & 30s & 28min & 10GB \\ 
    Flare   & - & - & -& - & - & -& - & - & - & - & - &- &- & - \\ 
    InstantSplat & 2.55 & 19.22 & 0.739 & 0.515 &58min &8min & 41GB & 2.23 & 13.09 & 0.414 & 0.680 &97min &7min & 40GB\\ 
    %\midrule
    \hline 
    \textbf{Ours}          & \textbf{1.41} & \textbf{31.53} & \textbf{0.915} & \textbf{0.245} &1min &63min & 14GB& \textbf{1.64} & \textbf{24.83} & \textbf{0.780} & \textbf{0.272} &8min &31min & 12GB\\ 
    \bottomrule
    \end{tabular}

    \label{table:main_result}
\end{table*}
%\vspace{-15pt}

\textbf{Metrics.} We evaluate camera pose estimation and scene reconstruction (in terms of image rendering quality). For pose, we report translation error via Absolute Trajectory Error (ATE), measured in meters (m). For reconstruction, we use PSNR, SSIM, and LPIPS. We also log training time and peak memory to assess efficiency and scalability.

\textbf{Baselines.} We compare our method with seven baselines, including COLMAP+SPSG \citep{schonberger2016sfm}, CF-3DGS \citep{fu2024colmap-free3dgs}, DUSt3R \citep{wang2024dust3r}, MASt3R \citep{leroy2024mast3r}, Fast3R \citep{Yang_2025_fast3R}, Flare \citep{zhang2025flare}, and  InstantSplat \citep{fan2024instantsplat}. Since COLMAP+SPSG, DUSt3R, MASt3R, and Fast3R only estimate point clouds and camera poses from images without directly producing Gaussians, we use the original 3DGS \citep{kerbl2023-3dgs} training pipeline for scene reconstruction, indicated with an asterisk (*).

\vspace{-7pt}
\begin{figure*}
  \centering
  \includegraphics[width=\textwidth]{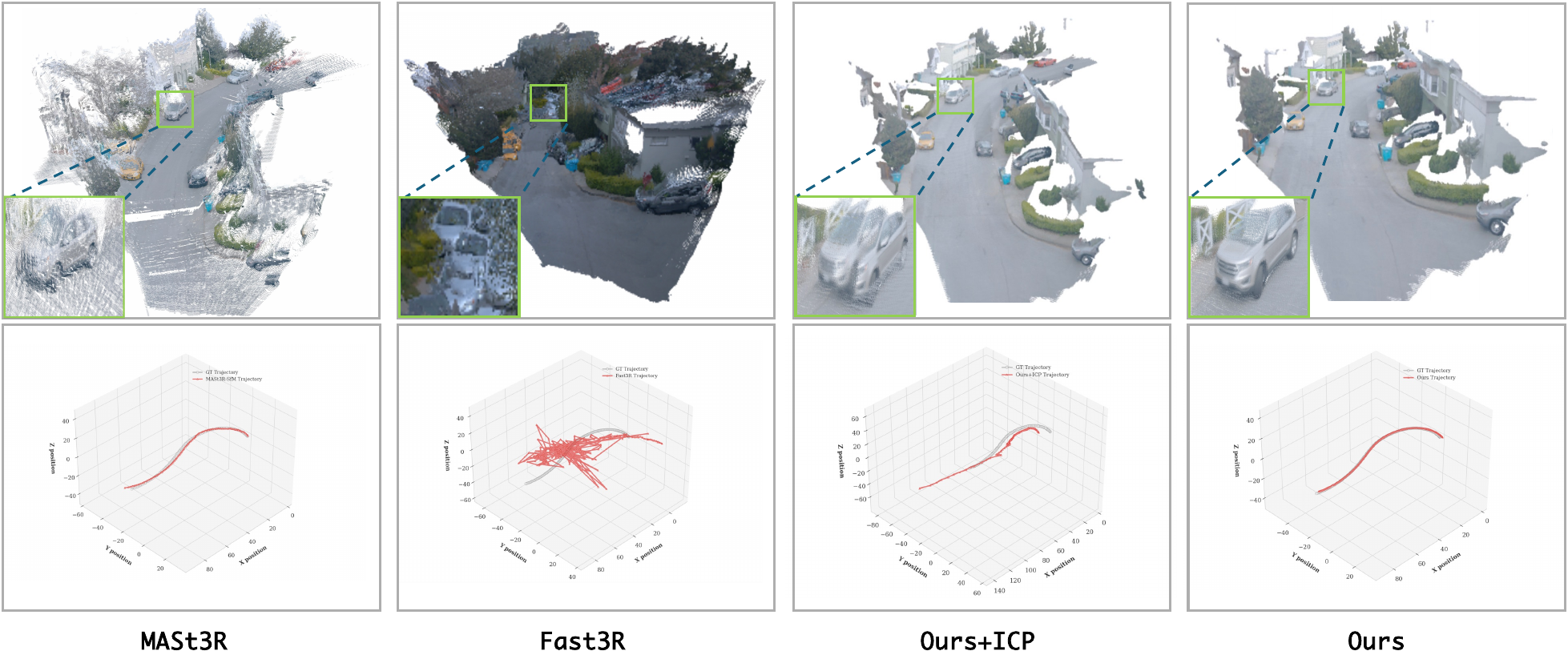}
  \vspace{-18pt}
  \caption{Qualitative comparison of reconstructed point clouds. The bottom row shows the estimated camera trajectories. Fast3R exhibits significant drift, while Ours+ICP still suffers from misalignment. Our method achieves accurate submap fusion and globally consistent pose estimation.}%In contrast, GlORIE-SLAM and OpenGS-SLAM exhibit rendering distortions and blurriness.
  \label{fig:point_cloud}
  \vspace{-15pt}
\end{figure*}

\subsection{Analysis of Experimental Results.}

We evaluate our method on the Waymo and KITTI datasets, with results summarized in Tab.~\ref{table:main_result}. InstantSplat, designed for sparse-view settings, fails to scale to large inputs due to memory limitations and performs poorly even when restricted to 80 views. Fast3R is efficient but suffers from severely inaccurate pose estimation. Flare supports only a limited number of input view sizes. COLMAP underperforms due to divergence in several scenes, where large errors skew the average ATE.

In contrast, our method combines the pretrained VGGT model, the PPM mapping module, and joint pose optimization to achieve superior reconstruction quality and trajectory accuracy. Fig.~\ref{fig:waymo_render} shows rendering results comparison in 6 scenes, including road layouts, building structures, vehicles, and vegetation. Fig.~\ref{fig:point_cloud} highlights the consistency of point clouds at submap boundaries. Compared to Fast3R, Mast3R, and ICP-based registration, our approach achieves seamless alignment across groups, with minimal drift in overlapping regions. The corresponding trajectory plots confirm the effectiveness of our pose refinement. Additionally, our method produces globally consistent and accurate point clouds and camera poses within just a few minutes.

\begin{figure*}
  \centering
  \includegraphics[width=0.98\textwidth]{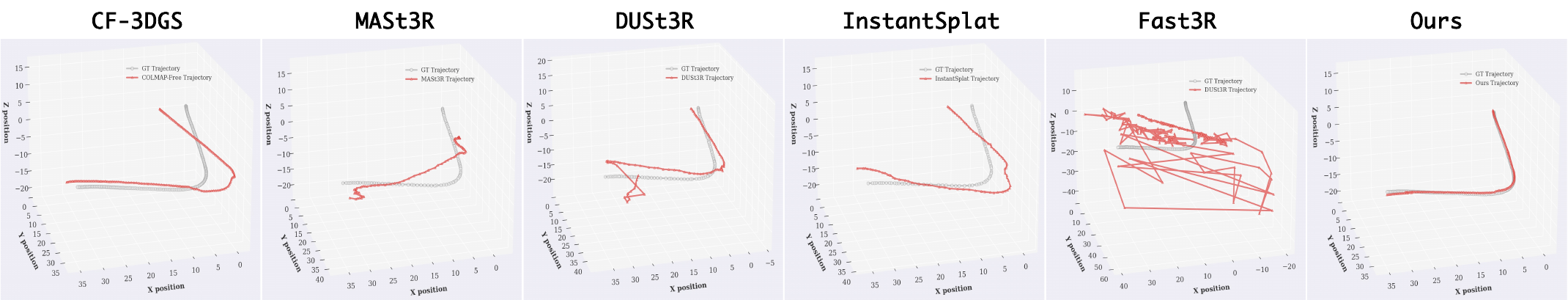}
  \caption{\textbf{Qualitative comparison of camera pose estimation.} Red denotes estimated poses while gray denotes ground truth. Our method achieves superior pose accuracy compared to other methods.}%In contrast, GlORIE-SLAM and OpenGS-SLAM exhibit rendering distortions and blurriness.
  \label{fig:ate}
  \vspace{-15pt}
\end{figure*}

Fig.~\ref{fig:ate} further demonstrates that our estimated trajectories are significantly more accurate and stable than those of competing methods. These results demonstrate the effectiveness of our framework for unposed reconstruction from hundreds of outdoor images.

\begin{wraptable}{r}{0.5\textwidth} 
\vspace{-12pt}
\centering
\caption{\textbf{Ablation results on the Waymo dataset.} Top: comparison of different submap alignment strategies based on our method. “Ours + ICP” and “Ours + COLMAP” denote using relative poses from ICP or COLMAP for submap mapping. Bottom: ablations of our Probabilistic Procrustes Mapping (PPM) and 3DGS pose joint optimization (Joint Opt.) modules.}
\vspace{-5pt}
\fontsize{8.5}{9.5}\selectfont  % 设置整体字体大小
% \setlength{\tabcolsep} % 控制列间距
% \begin{tabular}{>{\arraybackslash}p{3cm}|*{4}{>{\centering\arraybackslash}p{1.0cm}}*{2}{>{\centering\arraybackslash}p{1cm}}}

\begin{tabular}{@{}l|cccc@{}}
\toprule
Method & ATE & PSNR & SSIM & LPIPS  \\
\midrule
Ours w/ ICP    &3.24   &27.63 & 0.852  & 0.341 \\
Ours w/ COLMAP  &3.77   &28.13 & 0.835  & 0.336 \\
\midrule
w/o  PPM    &5.97   &27.54 & 0.824   & 0.363 \\
w/o   Joint Opt.   &0.68  &30.47  &0.903  &0.242\\
\midrule
\textbf{Ours}    &\textbf{0.56}  & \textbf{32.72}  &\textbf{0.935}  & \textbf{0.211}  \\
\bottomrule
\end{tabular}

\label{tab:ablation}
\vspace{-15pt}
\end{wraptable}

\subsection{Ablation Study}

We conduct ablation experiments on the Waymo dataset to evaluate the effectiveness of the proposed probabilistic Procrustes mapping (PPM) and the joint 3DGS optimization module. As shown in Table~\ref{tab:ablation}, we first compare different alignment strategies. Using ICP or COLMAP-predicted relative poses for submap registration leads to notable pose errors and visible misalignments in the final reconstructions. In contrast, our probabilistic Procrustes mapping module achieves significantly higher registration accuracy and reconstruction fidelity, demonstrating the advantage of combining closed-form alignment with probabilistic refinement.

We further ablate each core module to assess its individual impact. Disabling the PPM module and replacing it with VGGT relative pose estimation results in degraded global consistency and lower-quality novel view synthesis. Similarly, fixing camera poses during the 3DGS training stage leads to performance drops in both image quality and geometric coherence. These results confirm that jointly optimizing camera poses and scene representation is essential for accurate and robust reconstruction. Overall, both the PPM mapping module and joint pose optimization play critical roles in ensuring accurate global alignment and high-quality 3DGS reconstruction.

\subsection{Limitations}

Our approach relies on the quality of the pretrained MVS predictions for initial poses and geometry. While the joint optimization stage can correct moderate errors, severe inaccuracies in initialization may degrade the final reconstruction quality. As the number of input frames increases, accumulated drift and higher optimization costs can limit scalability to large-scale or long sequences. Moreover, in highly dynamic scenes with frequent motion or occlusions, the lack of consistent correspondences across views may hinder stable optimization and reduce reconstruction fidelity.

\section{Conclusion}
We presented a scalable and robust framework for unposed 3D Gaussian Splatting reconstruction. By integrating pretrained MVS models with a divide-and-conquer strategy, our method efficiently handles outdoor scenes with hundreds of uncalibrated views. We introduce a Probabilistic Procrustes Mapping module for global registration, followed by a 3DGS and poses joint optimization module for jointly refining camera poses and 3D Gaussians. Our method achieves state-of-the-art performance and offers practical value for unposed 3D reconstruction in real-world scenarios.

\newpage
\bibliographystyle{plainnat}
\bibliography{reference}

\newpage
\section*{Appendix: Quaternion–Point Jacobian}
\label{jacob1}
In this appendix, we derive the Jacobian of the rotated point \(R(q)\,\mathbf p\) with respect to the unit quaternion \(q = [w,\,v^\top]^\top\), where \(v = (x,y,z)^\top\).

We start from the equivalent expression for the rotation:
\begin{equation}
R(q)\,\mathbf p
= \bigl(w^2 - \|v\|^2\bigr)\,\mathbf p
+ 2\,v\,(v^\top \mathbf p)
+ 2\,w\,(v \times \mathbf p).
\end{equation}

Define:
\begin{equation}
\Delta(q) = R(q)\,\mathbf p = A + B + C,
\end{equation}
with:
\begin{equation}
A = (w^2 - v^\top v)\,\mathbf p,\quad
B = 2\,v\,(v^\top \mathbf p),\quad
C = 2\,w\,(v \times \mathbf p).
\end{equation}

\subsection*{1. Derivative with respect to \(w\)}

Only \(A\) and \(C\) depend on \(w\). We have:
\begin{equation}
\frac{\partial A}{\partial w}
= 2w\,\mathbf p,\qquad
\frac{\partial C}{\partial w}
= 2\,(v \times \mathbf p).
\end{equation}

Therefore:
\begin{equation}
\frac{\partial\,(R(q)\,\mathbf p)}{\partial w}
= 2\,w\,\mathbf p + 2\,(v \times \mathbf p).
\end{equation}

\subsection*{2. Derivative with respect to \(v\)}

Let \([\mathbf p]_\times\) denote the skew‐symmetric matrix such that \([\mathbf p]_\times u = \mathbf p \times u\).  We compute:
\begin{equation}
\frac{\partial A}{\partial v}
= -2\,(v^\top \mathbf p)\,I_{3}
= -2\,\mathbf p\,v^\top,
\end{equation}

\begin{equation}
\frac{\partial B}{\partial v}
= 2\,(v^\top \mathbf p)\,I_{3}
+ 2\,v\,\mathbf p^\top,
\end{equation}

\begin{equation}
\frac{\partial C}{\partial v}
= 2\,w\,[\mathbf p]_\times.
\end{equation}

Combining these terms yields:
\begin{equation}
\label{QPJ}
\frac{\partial\,(R(q)\,\mathbf p)}{\partial v}
= -2\,\mathbf p\,v^\top
+ 2\,(v^\top \mathbf p)\,I_{3}
+ 2\,v\,\mathbf p^\top
+ 2\,w\,[\mathbf p]_\times,
\end{equation}

\subsection*{3. Assembling the \(3\times4\) Jacobian}

Stacking the partial derivatives with respect to \(w\) and \(v\) produces the full Jacobian:
\begin{equation}
\frac{\partial\,(R(q)\,\mathbf p)}{\partial q}
= \biggl[\,
\underbrace{2w\,\mathbf p + 2\,(v\times\mathbf p)}_{3\times1}
\;\Bigm|\;
\underbrace{
-2\,\mathbf p\,v^\top
+2\,(v^\top \mathbf p)\,I_{3}
+2\,v\,\mathbf p^\top
+2\,w\,[\mathbf p]_\times
}_{3\times3}
\biggr]_{3\times4}\,.
\end{equation}
where the first column corresponds to \(\partial/\partial w\) and the remaining three columns correspond to \(\partial/\partial x,\partial/\partial y,\partial/\partial z\). This Jacobian can be directly used in gradient‐based optimization.

\clearpage

\section*{Appendix: Analytic Gradients of \(\mu'\) and \(\Sigma'\) w.r.t.\ Pose \(T\)}
\label{jacob2}
In this appendix, we derive the analytic gradients of the projected coordinate \(\mu' = \pi(\mu_c)\) and the projected covariance
\(\Sigma' = J\,R(q)\,\Sigma R(q)^\top J^\top\)
with respect to the camera pose \(T=(R(q),\,t)\), where
\[
\mu_c = R(q)\,\mu + t,\quad
q=[q_r,q_i,q_j,q_k]^\top,\quad
t\in\mathbb R^3,
\]
\(\mu\in\mathbb R^3\) is a 3D point, and \(J=\frac{\partial\pi(\mu_c)}{\partial\mu_c}\) is the \(2\times3\) projection Jacobian.

\subsection*{1. Gradients of the projected coordinate \(\mu'\)}

By chain rule, the derivative w.r.t.\ translation is
\begin{equation}
\frac{\partial \mu'}{\partial t}
= J \,\frac{\partial \mu_c}{\partial t}
= J \;=\;
\begin{bmatrix}
\displaystyle \frac{f_x}{z_c} & 0 & -\displaystyle\frac{f_x\,x_c}{z_c^2}\\[8pt]
0 & \displaystyle\frac{f_y}{z_c} & -\displaystyle\frac{f_y\,y_c}{z_c^2}
\end{bmatrix}.
\end{equation}
The derivative w.r.t.\ the quaternion is
\begin{equation}
\frac{\partial \mu'}{\partial q}
= J \,\frac{\partial \mu_c}{\partial q}
= J 
\begin{bmatrix}
\frac{\partial \mu_c}{\partial q_r} & \frac{\partial \mu_c}{\partial q_i}
& \frac{\partial \mu_c}{\partial q_j} & \frac{\partial \mu_c}{\partial q_k}
\end{bmatrix}.
\end{equation}

Here the \(3\times1\) blocks \(\partial\mu_c/\partial q_\alpha\) are:

\begin{equation}
\frac{\partial \mu_c}{\partial q_r}
= 2
\begin{bmatrix}
0      & -q_k   &  q_j \\
q_k    &  0     & -q_i \\
-q_j   &  q_i   &  0
\end{bmatrix}\mu,
\end{equation}
\begin{equation}
\frac{\partial \mu_c}{\partial q_i}
= 2
\begin{bmatrix}
0      &  q_j   &  q_k \\
q_j    & -2q_i  & -q_r \\
q_k    &  q_r   & -2q_i
\end{bmatrix}\mu,
\end{equation}
\begin{equation}
\frac{\partial \mu_c}{\partial q_j}
= 2
\begin{bmatrix}
-2q_j  &  q_i   &  q_r \\
q_i    &  0     &  q_k \\
q_r    &  q_k   & -2q_j
\end{bmatrix}\mu,
\end{equation}
\begin{equation}
\frac{\partial \mu_c}{\partial q_k}
= 2
\begin{bmatrix}
-2q_k  & -q_r   &  q_i \\
q_r    & -2q_k  &  q_j \\
q_i    &  q_j   &  0
\end{bmatrix}\mu.
\end{equation}

\subsection*{2. Gradients of the projected covariance \(\Sigma'\)}

Since translation does not affect covariance:
\begin{equation}
\frac{\partial \Sigma'}{\partial t}
= 0.
\end{equation}

For the quaternion:
\begin{equation}
\frac{\partial \Sigma'}{\partial q}
= J\,\frac{\partial \bigl(R\,\Sigma_{\mathrm w}\,R^\top\bigr)}{\partial q}\,J^\top
= J\Bigl(\frac{\partial R}{\partial q}\,\Sigma_{\mathrm w}\,R^\top
+ R\,\Sigma_{\mathrm w}\,\frac{\partial R^\top}{\partial q}\Bigr)J^\top,
\end{equation}
where \(\partial R/\partial q\) is the classic gradient of the rotation matrix with respect to the quaternion, and \(\partial R^\top/\partial q\) is its transpose.

These closed-form derivatives enable efficient back-propagation of both \(\mu'\) and \(\Sigma'\) through the differentiable 3DGS rendering pipeline.

\clearpage
\section*{NeurIPS Paper Checklist}

\begin{enumerate}

\item {\bf Claims}
    \item[] Question: Do the main claims made in the abstract and introduction accurately reflect the paper's contributions and scope?
    \item[] Answer: \answerYes{} % Replace by \answerYes{}, \answerNo{}, or \answerNA{}.
    \item[] Justification: {The abstract and introduction accurately summarize the contributions of the paper, including the proposed method, key technical insights, and empirical improvements. The claims are well-supported by theoretical analysis and experimental results.}

\item {\bf Limitations}
    \item[] Question: Does the paper discuss the limitations of the work performed by the authors?
    \item[] Answer: \answerYes{} % Replace by \answerYes{}, \answerNo{}, or \answerNA{}.
    \item[] Justification: {The paper includes a dedicated Limitations section discussing assumptions, potential failure cases, and generalizability issues.}

\item {\bf Theory assumptions and proofs}
    \item[] Question: For each theoretical result, does the paper provide the full set of assumptions and a complete (and correct) proof?
    \item[] Answer: \answerNA{} % Replace by \answerYes{}, \answerNo{}, or \answerNA{}.
    \item[] Justification:  {The paper does not contain formal theoretical results or proofs.}

    \item {\bf Experimental result reproducibility}
    \item[] Question: Does the paper fully disclose all the information needed to reproduce the main experimental results of the paper to the extent that it affects the main claims and/or conclusions of the paper (regardless of whether the code and data are provided or not)?
    \item[] Answer: \answerYes{} % Replace by \answerYes{}, \answerNo{}, or \answerNA{}.
    \item[] Justification:  {The paper provides complete details about the experimental setup, dataset usage, model architecture, training procedures, and evaluation protocols. Sufficient information is included to enable reproduction of all key results.}

\item {\bf Open access to data and code}
    \item[] Question: Does the paper provide open access to the data and code, with sufficient instructions to faithfully reproduce the main experimental results, as described in supplemental material?
    \item[] Answer: \answerYes{} % Replace by \answerYes{}, \answerNo{}, or \answerNA{}.
    \item[] Justification:  {The code and data will be released upon acceptance, with complete instructions for reproducing the main results.}

\item {\bf Experimental setting/details}
    \item[] Question: Does the paper specify all the training and test details (e.g., data splits, hyperparameters, how they were chosen, type of optimizer, etc.) necessary to understand the results?
    \item[] Answer: \answerYes{} % Replace by \answerYes{}, \answerNo{}, or \answerNA{}.
    \item[] Justification:  {We report experimental details in our main paper.}

\item {\bf Experiment statistical significance}
    \item[] Question: Does the paper report error bars suitably and correctly defined or other appropriate information about the statistical significance of the experiments?
    \item[] Answer: \answerNo{} % Replace by \answerYes{}, \answerNo{}, or \answerNA{}.
    \item[] Justification:  {The paper reports PSNR, SSIM, ATE, LPIPS, which are commonly used as a measure of performance in image processing experiments. This approach is standard in the field and is sufficient to convey the performance of the methods under investigation. }

\item {\bf Experiments compute resources}
    \item[] Question: For each experiment, does the paper provide sufficient information on the computer resources (type of compute workers, memory, time of execution) needed to reproduce the experiments?
    \item[] Answer: \answerYes{} % Replace by \answerYes{}, \answerNo{}, or \answerNA{}.
    \item[] Justification:  {We describe the computing environment used in our experiments, including GPU types, memory size, number of training hours.}

    
\item {\bf Code of ethics}
    \item[] Question: Does the research conducted in the paper conform, in every respect, with the NeurIPS Code of Ethics \url{https://neurips.cc/public/EthicsGuidelines}?
    \item[] Answer: \answerYes{} % Replace by \answerYes{}, \answerNo{}, or \answerNA{}.
    \item[] Justification:  { We do not foresee any ethical concerns related to data usage, environmental impact, or fairness.}

\item {\bf Broader impacts}
    \item[] Question: Does the paper discuss both potential positive societal impacts and negative societal impacts of the work performed?
    \item[] Answer: \answerYes{} % Replace by \answerYes{}, \answerNo{}, or \answerNA{}.
    \item[] Justification:  {The paper includes a Broader Impact section discussing potential societal applications of our 3D scene reconstruction framework.}
    
\item {\bf Safeguards}
    \item[] Question: Does the paper describe safeguards that have been put in place for responsible release of data or models that have a high risk for misuse (e.g., pretrained language models, image generators, or scraped datasets)?
    \item[] Answer: \answerNA{} % Replace by \answerYes{}, \answerNo{}, or \answerNA{}.
    \item[] Justification:  {The paper does not release models or data with high misuse risk}

\item {\bf Licenses for existing assets}
    \item[] Question: Are the creators or original owners of assets (e.g., code, data, models), used in the paper, properly credited and are the license and terms of use explicitly mentioned and properly respected?
    \item[] Answer: \answerYes{} % Replace by \answerYes{}, \answerNo{}, or \answerNA{}.
    \item[] Justification:  {All third-party datasets and tools used in the paper are properly cited with licenses stated where applicable.}

\item {\bf New assets}
    \item[] Question: Are new assets introduced in the paper well documented and is the documentation provided alongside the assets?
    \item[] Answer: \answerNA{} % Replace by \answerYes{}, \answerNo{}, or \answerNA{}.
    \item[] Justification:  {The paper does not introduce new datasets or models requiring documentation.}

\item {\bf Crowdsourcing and research with human subjects}
    \item[] Question: For crowdsourcing experiments and research with human subjects, does the paper include the full text of instructions given to participants and screenshots, if applicable, as well as details about compensation (if any)? 
    \item[] Answer: \answerNA{} % Replace by \answerYes{}, \answerNo{}, or \answerNA{}.
    \item[] Justification:  {The research does not involve crowdsourcing or experiments with human subjects.}

\item {\bf Institutional review board (IRB) approvals or equivalent for research with human subjects}
    \item[] Question: Does the paper describe potential risks incurred by study participants, whether such risks were disclosed to the subjects, and whether Institutional Review Board (IRB) approvals (or an equivalent approval/review based on the requirements of your country or institution) were obtained?
    \item[] Answer: \answerNA{} % Replace by \answerYes{}, \answerNo{}, or \answerNA{}.
    \item[] Justification:   {No human subjects were involved in the research, so IRB approval is not applicable.}

\item {\bf Declaration of LLM usage}
    \item[] Question: Does the paper describe the usage of LLMs if it is an important, original, or non-standard component of the core methods in this research? Note that if the LLM is used only for writing, editing, or formatting purposes and does not impact the core methodology, scientific rigorousness, or originality of the research, declaration is not required.
    %this research? 
    \item[] Answer: \answerNA{} % Replace by \answerYes{}, \answerNo{}, or \answerNA{}.
    \item[] Justification:  {LLMs were not used in the design or implementation of the core methods in the paper. They were only used for minor editing support.}

\end{enumerate}

\end{document}